\documentclass[10pt, a4paper]{article}

\usepackage[final]{lrec-coling2024} 
\usepackage[inkscapelatex=false]{svg} 
\usepackage{color} 
\usepackage{booktabs} 
\usepackage{diagbox} 
\usepackage{times}
\usepackage{latexsym}
\usepackage{multirow}
\usepackage{graphicx}
\usepackage{comment}
\usepackage{CJKutf8}
\usepackage[utf8]{inputenc}
\usepackage[justification=centering]{caption}
\usepackage{amsmath}

\usepackage{times}
\usepackage{latexsym}

\title{
An Empirical Study on the Robustness of Massively Multilingual Neural Machine Translation}

\name{Supryadi, Leiyu Pan, Deyi Xiong$^{\ast}$\thanks{$^{\ast}$Corresponding author.}} 

\address{College of Intelligence and Computing, Tianjin University \\
         Tianjin, China \\
         \{supryadi, lypan, dyxiong\}@tju.edu.cn\\}

\abstract{
Massively multilingual neural machine translation (MMNMT) has been proven to enhance the translation quality of low-resource languages. In this paper, we empirically investigate the translation robustness of Indonesian-Chinese translation in the face of various naturally occurring noise. To assess this, we create a robustness evaluation benchmark dataset for Indonesian-Chinese translation. This dataset is automatically translated into Chinese using four NLLB-200 models of different sizes. We conduct both automatic and human evaluations. Our in-depth analysis reveal the correlations between translation error types and the types of noise present, how these correlations change across different model sizes, and the relationships between automatic evaluation indicators and human evaluation indicators. The dataset is publicly available at \url{https://github.com/tjunlp-lab/ID-ZH-MTRobustEval}.
 \\ \newline  }

\begin{document}

\maketitleabstract

\section{Introduction}

Recent years have witnessed that neural machine translation (NMT) achieves a remarkable progress in both high- and low-resource language translation. For the former aspect, translation quality is substantially improved for many high-resource language pairs (e.g., Chinese-English, French-English) over the years, which has been tracked by yearly WMT evaluation \cite{bojar-etal-2018-findings}. Human parity has even been reached for some language pairs in terms of certain evaluation protocols (\citealp{human-parity,barrault-etal-2019-findings}). For the latter aspect, to improve translation quality of low-resource languages, massively multilingual neural machine translation (MMNMT) has been explored with growing interest, which enables knowledge transfer from high-resource languages to low-resource languages (\citealp{aharoni-etal-2019-massively,11125,Team2022}).

Conversely, NMT still faces challenges related to robustness, particularly in handling noise \citep{belinkov2018synthetic} and adapting to domain shifts \citep{lai-etal-2022-improving-domain}. In this study, we aim to delve into the translation robustness of Indonesian-Chinese within the context of massively multilingual NMT. Our specific objectives include understanding: 1) the patterns of the relationship between translation error types and noise types, and 2) how these patterns change across various MMNMT model sizes, ranging from models with millions to billions of parameters. Such an investigation holds significant importance in advancing our understanding of the robustness of Indonesian-Chinese translation, which remains an underexplored area, and in the development of MMNMT models.

To empirically study these patterns and relations, we use the open-sourced NLLB-200 \cite{Team2022} models as our MMNMT models. We curate an Indonesian-to-Chinese translation robustness evaluation dataset that consists of 1001 sentence pairs. Both languages are among the top-20 most spoken languages in the world but the parallel resources for them are very limited. We crawl noisy Indonesian sentences from social medias and manually translate them into Chinese with the collaboration between source language local speaker and the expert of the target language.

We manually identify noises in the source language and categorize them into 10 groups. These noisy source sentences are then automatically translated into Chinese with four NLLB-200 models of different sizes, where translation errors in translated target sentences are detected and classified into 10 categories. In addition to automatic evaluation of translation results with BLEU \cite{papineni-etal-2002-bleu} and CHRF++ \cite{popovic-2017-chrf}, we also conduct human evaluation with multidimensional quality metric (MQM\footnote{\url{www.themqm.org}}).

\begin{figure*}[t]
    \centering
    \includegraphics[scale=0.43]{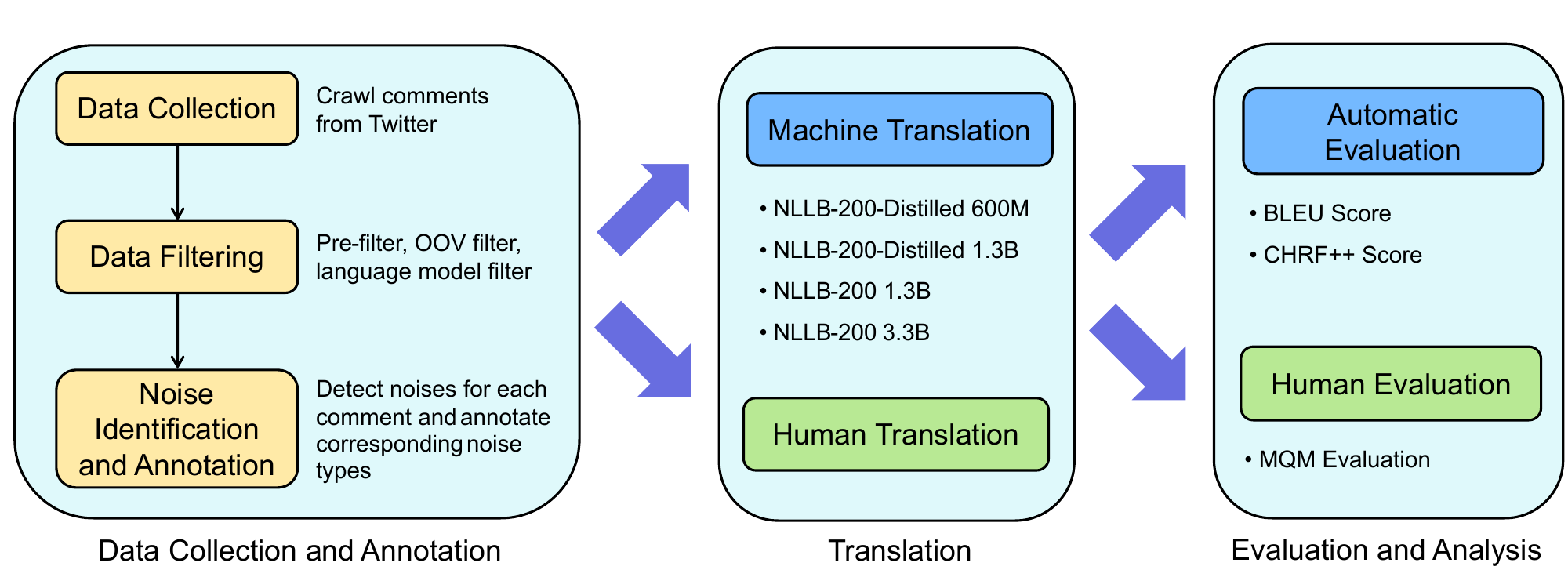}
    \caption{Robustness evaluation and analysis protocol.}
    \label{fig:0}
\end{figure*}

The contributions of our work are as follows:
\begin{itemize}
  \item We empirically evaluate the robustness for Indonesian-Chinese translation. To the best of our knowledge, this is the first attempt to study the robustness of Indonesian-Chinese translation based on MMNMT models. 
  \item We curate a new noisy parallel dataset on Indonesian-Chinese translation for such evaluation. 
  \item We manually identify noise types in the dataset and translation error types in translations generated by the NLLB-200 models, study the relation patterns of them and examine the changes of these patterns across different model sizes. 
\end{itemize}

\section{Related Work}
\paragraph{NMT Robustness and Evaluation} Robustness is of paramount importance for neural machine translation (NMT), especially when NMT systems are deployed in real-world applications. A wide variety of efforts have been dedicated to enhancing the robustness of NMT. Among them, black-box methods are widely explored \cite{10.1007/978-3-319-64206-2_27,Karpukhin2019,liu-etal-2019-robust, wallace-etal-2020-imitation,9413586,zhang-etal-2021-crafting}. Alternatively, white-box methods, employing gradient-based approaches, have also been proposed \cite{Cheng2019}. Moreover, empirical evidence suggests that attacking NMT from the source side yields greater effectiveness \cite{zeng-xiong-2021-empirical}. These methods usually employ synthetic noise to improve robustness. 

In the context of robustness towards natural noise, the MTNT dataset \citep{michel-neubig-2018-mtnt} is designed, originating from noisy data collected from Reddit\footnote{\url{www.reddit.com}} comments. This dataset comprises three different languages: English, French, and Japanese. In a similar vein, for the assessment of French-English translation robustness, noisy data have also been gathered from restaurant reviews \citep{restaurant}. Additionally, for the evaluation of Chinese-English translation robustness, a dialogue dataset has been created as the natural noise data \cite{wang-etal-2021-autocorrect}.

Partially inspired by \citet{michel-neubig-2018-mtnt}, we have curated a novel robustness evaluation dataset. However, our dataset differs significantly from \citet{michel-neubig-2018-mtnt} in three aspects. Firstly, our primary focus lies in assessing the robustness of Indonesian-Chinese translation from two geographically distant languages. Secondly, our noisy data is derived from Twitter comments rather than Reddit, encompassing a broader spectrum of topics. Thirdly, we have manually identified and annotated different noise types for each sentence pair, enabling a more targeted evaluation of noise-specific robustness.

\paragraph{Multilingual NMT} Multilingual neural machine translation (MNMT) has garnered growing interest in recent years owing to its capacity to facilitate the deployment of NMT systems supporting multiple languages, knowledge transfer between languages \cite{sun-xiong-2022-language}, and zero-shot translation capabilities, among others \cite{xu-etal-2021-modeling,li-etal-2023-mmnmt}. To enable knowledge transfer across an extensive array of languages, including 100 or more languages, research has delved into massively multilingual neural machine translation (MMNMT) \cite{10.1162/tacl_a_00065,jin-xiong-2022-informative}. This exploration has evolved from English-centric models \cite{aharoni-etal-2019-massively} to models extending beyond English-centric approaches, such as M2M-100 \cite{11125}.

Among those non-English-centric models, NLLB-200 \cite{Team2022} has recently been open-sourced, which encompasses 200 languages and 40,000 translation directions, supported by a model with up to 54 billion parameters trained on a huge amount of natural and synthesized data. In this study, we employ NLLB-200 models to assess the robustness of MMNMT on non-English languages using our curated dataset. 

\paragraph{Multilingual NMT Robustness}
The robustness of multilingual NMT also been evaluated recently \cite{pan-etal-2023-robustness}. A variety of noises at the character-, word-, and multiple levels have been explored for the study of multilingual NMT robustness. It has been observed that the robustness of multilingual NMT can be transferred across languages. In contrast to previous study, this research specifically focuses evaluating the robustness of MMNMT towards naturally occurring noise, which is categorized into 10 distinct types.

\section{Robustness and Evaluation Protocol}

We propose a general protocol for evaluating and analyzing MMNMT robustness towards naturally occurring noises, which is model- and language-independent. As illustrated in Figure~\ref{fig:0}, the protocol consists of three main stages. 

\begin{itemize}
  \item [1)] 
  \textbf{Data Collection and Annotation}: In this initial stage, we commence by identifying suitable sources for collecting noisy data. Once the sources are determined, data is extracted from these sources. We employ automatic noise detection methods to filter the extracted data, retaining only the noisy portions. In our study, this extraction and detection process yields a high-quality monolingual Indonesian corpus that incorporates naturally occurring noise. Each sentence in this corpus is then labeled with its associated noise category. It is worth noting that each sentence may be annotated with multiple noise categories.
  \item [2)]
  \textbf{Translation}: The collected source corpus is then translated into the target language by human translators, adhering to a noise translation convention to ensure consistency in the translation of noisy fragments throughout the entire corpus. These manual translations serve as reference translations for both automatic evaluation and manual analysis. To assess and analyze the robustness of specific MMNMT models, the source corpus is also automatically translated into the target language by these MMNMT models.
  \item [3)]
  \textbf{Evaluation and Analysis}: In this stage, we carry out both automatic and human evaluations. We manually identify translation errors and categorize them according to multidimensional quality metrics (MQM) (specifically level-1 error types). With annotated noise types and translation error types, we can conduct a thorough and comprehensive analysis of the robustness issues observed in MMNMT models.
\end{itemize}

We will detail the data collection and annotation procedure in Section \ref{section:4}, translation in Section \ref{section:5} and evaluation in Section \ref{section:6}. In-depth analysis results are presented in Section \ref{section:7}.

\begin{table*}[t]
    \centering
    \resizebox{\textwidth}{!}{
    \begin{tabular}{l|cccccccccc}
    \bottomrule
        \multicolumn{1}{c|}{\textbf{Indonesia Text}} & \textcolor[RGB]{231, 76, 60}{\textbf{Spelling/Typo}} & \textcolor[RGB]{230, 126, 34}{\textbf{Grammar}} & \textcolor[RGB]{255, 159, 243}{\textbf{Spoken Language}} & \textcolor[RGB]{123, 237, 159}{\textbf{Slang}} & \textcolor[RGB]{39, 174, 96}{\textbf{Proper Noun}} & \textcolor[RGB]{72, 219, 251}{\textbf{Dialect}} & \textcolor[RGB]{46, 134, 222}{\textbf{Code Switching}} & \textcolor[RGB]{41, 128, 185}{\textbf{Jargon}} & \textcolor[RGB]{155, 89, 182}{\textbf{Emojis}} & \textcolor[RGB]{211, 84, 0}{\textbf{Slurs}}  \\ \hline
        Makasih \textcolor[RGB]{72, 219, 251}{kang} sial. Berkat \textcolor[RGB]{46, 134, 222}{lu} dukung \textcolor[RGB]{39, 174, 96}{prancis}\\ argentina jadi juara. Sekali lagi terima kasih sudah \textcolor[RGB]{255, 159, 243}{bikin}\\ prancis sial
        & ~ & ~ & V & ~ & ~ & V & V & ~ & V & \\ \hline
        \textcolor[RGB]{230, 126, 34}{alah} ribet amat urusan ucap natal \textcolor[RGB]{123, 237, 159}{ga} ucap \textcolor[RGB]{231, 76, 60}{nata},\\ noh negara lain udah \textcolor[RGB]{255, 159, 243}{mikirin} hidup di \textcolor[RGB]{39, 174, 96}{mars}. & V & V & V & V & V & ~ & ~ & ~ & ~ &   \\ \hline
        Gara2 \textcolor[RGB]{255, 159, 243}{ngomen} denny Akun\textcolor[RGB]{230, 126, 34}{.ku} dihanguskan \textcolor[RGB]{211, 84, 0}{njir} & ~ & V & V & ~ & ~ & ~ & ~ & V & ~ & V  \\
    \bottomrule
    \end{tabular}
    }
\caption{\label{table-1}
Noise identification and annotation.
}
\end{table*}

\section{Data Collection and Annotation}
\label{section:4}
\subsection{Data Collection}

We collected raw social media comments from Twitter. To obtain these comments, we utilized Tweepy\footnote{\url{https://github.com/tweepy}}, a Python library for accessing the Twitter API. Given our focus on the Indonesian language, the comments are crawled using popular Twitter accounts from Indonesia as keywords. The collection period for these comments spanned one week, from 13 December 2022 to 20 December 2022. The final dataset contains a total of 25,973 comments.

\subsection{Data Filtering}

After the collection of these comments, we employed filtering methods to detect noisy comments, following the approach outlined by \citet{michel-neubig-2018-mtnt} in the MTNT dataset. We utilized three filtering methods for this purpose.

\paragraph{Pre-filter} We performed a pre-filtering process on the collected raw data in three steps, with the aim of retaining only naturally occurring noises: 
\begin{itemize}
  \item [1)] 
  Removing comments containing URLs.
  \item [2)]
  Removing comments from users where their usernames contain “bot” or “AutoModerator”.
  \item [3)]
  Removing comments written in other languages. We use Python library Langid.py\footnote{\url{https://github.com/saffsd/langid.py}} to detect non-Indonesian languages.
\end{itemize}

\paragraph{OOV Filter} For robustness evaluation, we aimed to make the corpus as noisy as possible, considering out-of-vocabulary (OOV) words as a form of noise. To introduce unknown words and add noise to the sentences, we created a dictionary using a contrast corpus. Our contrast corpus comprises the Indonesian section of WMT20news-commentary-v15\footnote{\url{https://www.statmt.org/wmt20}} and OpenSubtitles\footnote{\url{https://www.opensubtitles.com}}. Using the fairseq tool \cite{ott2019fairseq}, we generate a dictionary containing 5,000 words. Then, we only keep those comments that contain at least one OOV word.

\paragraph{Language Model Filter} In the final step, we employed an n-gram language model to further identify noisy comments. We tokenized both the contrast corpus and the collected comments using Byte-Pair Encoding (BPE) with SentencePiece \cite{kudo-richardson-2018-sentencepiece}. We then trained a 5-gram Kneser-Ney smoothed language model on the segmented contrast corpus. This trained model is used to calculate language model scores for the comments, normalized by sentence length. We selected comments within a specific score interval, specifically the range between the first and third quartiles of normalized language model scores in the comment corpus. However, we ensured that the normalized language model score of a retained comment was smaller than the third quartile of normalized language model scores in the contrast corpus. This approach strikes a balance, ensuring that each kept comment contains a certain amount of noise without being overly noisy, which differs from the method used in MTNT.

After applying the three filters to the collected raw data, we retained 1,001 comments in the final dataset. The average sentence length of these retained comments is 10 words, with a standard deviation of 6.6. The shortest comment consists of 1 word, while the longest comment comprises 48 words.

\begin{figure}[t]
    \centering
    \includegraphics[scale=0.38]{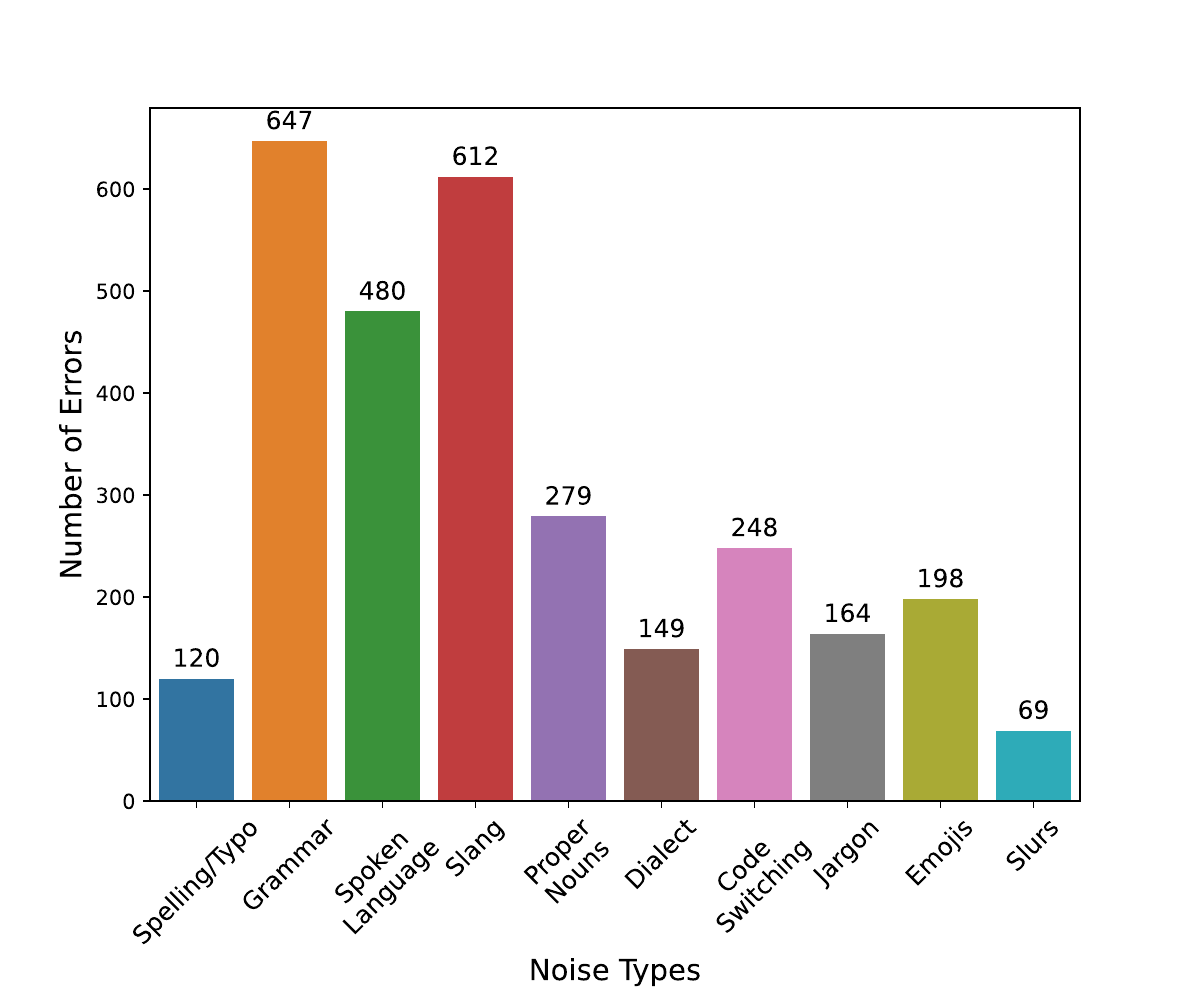}
    \caption{Statistics of different noise types in the curated dataset.}
    \label{fig:1}
\end{figure}

\subsection{Noise Identification and Annotation}
 We use a noise taxonomy similar to that used in MTNT, which consists of:

 \begin{itemize}
  \item \textbf{Spelling/typographical errors}: Comments contain incorrectly spelled or typed words.
  \item \textbf{Grammatical errors}: Comments are not grammatically written.
  \item \textbf{Spoken language}: Comments are written in the style of spoken language.
  \item \textbf{Internet slang}: Comments contain trending words in internet/social media.
  \item \textbf{Proper nouns}: Proper nouns, e.g., entities of place, person, are incorrectly written.
  \item \textbf{Dialects}: Comments contain Indonesia dialects, e.g., Javanese, Acehnese, and Balinese.
  \item \textbf{Code switching}: Comments contain more than one language.
  \item \textbf{Jargon}: Comments include specific words used in certain areas of life (environment).
  \item \textbf{Emojis}: Comments contain emojis for feeling expression.
  \item \textbf{Slurs}: Improper words are used to insult people.
\end{itemize}

In the corpus, we detected each instance of noise and classified them into their respective noise types, as previously described. It is worth emphasizing that a sentence may contain multiple types of noise, and we annotate each of these noise types for the sentence accordingly.

The example of our noise identifications and annotations is presented in Table \ref{table-1}. In the first sentence, ``kan'' is a Sundanese dialect with the meaning ``brother'', and ``lu'' is a Hokkien dialect meaning ``you''. These two words indicate the presence of code switching in the sentence. Additionally, the word ``prancis'' is a proper noun referring to ``France'', which should be capitalized as ``Prancis''. Towards the end of the sentence, the spoken word ``bikin'' should be replaced with the written word ``buat'', and an emoticon is present at the end of the sentence.

In the second sentence, the first word ``alah'' is an interjection expressing ``complaining''. There is a missing conjunction ``and'' between ``ucap natal'' and ``ga ucap natal''. ``Ga'' is a slang term that translates to ``no'' in English. The word ``nata'' appears to be a typographical error and should be corrected to ``Natal'', which means ``Christmas''. Similarly, ``mikirin'' is a spoken language form and should be written as ``memikirkan'', which means ``thinking''. Finally, ``Mars'' is a proper noun referring to the name of a planet.

\begin{table*}[t]
\centering
\resizebox{\textwidth}{!}{%
\begin{tabular}{ll|l}
\bottomrule
\multicolumn{1}{l}{\textbf{Error Type}} & ~&\multicolumn{1}{c}{\textbf{Description}}                                                      \\ \hline

\multirow{1}{*}{Terminology}     & ~     & Inconsistency and accuracy issues of the terminology.\\ \hline
\multirow{4}{*}{Accuracy} & Mistranslation  & Target content that does not accurately represent the source content.\\     
                                         & Omission           & The target content is missing from the translation that is present in the source. \\   
                                         & Addition & Target content that includes content not present in the source.          \\
                                         & Untranslated      & The text in source content is left untranslated in the target content.   \\
                                         & Hallucination      & The translation is very different or irrevelant with the source.   \\
\hline \multirow{2}{*}{Fluency}      & Grammar     & The translation result violates the grammatical rules of the target language.          \\ 
                                         & Punctuation           & Incorrect punctuation  for the locale or style.                   \\
                                        
\hline \multirow{1}{*}{Local Convention}     & ~     & The translation violates locale-specific content or formatting requirements.  \\
\hline \multirow{1}{*}{Audience Appropriateness}     & ~     & The use of content in the translation that is invalid or inappropriate for the target audience.  \\
\bottomrule
\end{tabular}%
}
\caption{MQM hierarchy.}
\label{table-2}
\end{table*}

\begin{table*}[]
\resizebox{\textwidth}{!}{%
\begin{tabular}{c|l}
\hline
\multicolumn{1}{c|}{\textbf{Severity Level}} & \multicolumn{1}{c}{\textbf{Description}}                                                      \\ \hline
Critical & The errors that significantly affect translation usability, understandability, and meaning. \\ 
Major                                  & Errors that would impact usability or understandability of the   translation.         \\ 
Minor                                  & Errors that would not impact the usability or understandability of the   translation. \\ \hline
\end{tabular}%
}
\caption{MQM severity levels.}
\label{table-3}
\end{table*}

In the last sentence, ``ngomen'' is a spoken language form that should be written as ``mengomentari'', which means ``to give comments''. The use of ``.k'' is a grammar error since the sentence is not finished, but a full stop is placed in the middle. Furthermore, it is important to note that ``njir'' is a derogatory slang term in the Indonesian language that dehumanizes individuals by comparing them to dogs.

The statistics for annotated noise types are presented in Figure \ref{fig:1}. The most prevalent noise type is grammatical noise, which is observed in 647 comments. Additionally, the spoken language and slang noise types are also prominently represented. This observation aligns well with our expectations for social media texts.

\section{Translation}
\label{section:5}

The annotated corpus is then translated into Chinese manually, creating a parallel corpus that acts as a benchmark testbed for evaluating robustness. We enlisted language experts proficient in both the source and target languages for this task. Through collaboration with experts in both languages, the sentences' meanings are conveyed with greater accuracy, making the translations more comprehensible to the reader. These translations were reviewed and proofread by professional translator to ensure translation consistency, particularly in noisy parts of the corpus. To curate a high-quality parallel corpus, it is important to establish guidelines ensuring consistency in the translation results. Our guidelines for the parallel corpus translation include:
\begin{itemize}
    \item Translating punctuation according to the target language convention and normalizing its usage.
    \item Translating idioms to convey their meaning and ensure reader comprehension.
    \item Correcting grammar errors in the source sentences during translation, maintaining proper grammar.
    \item Standardizing the translation of proper nouns (names of people, places, organizations, products) across the entire corpus.
    \item Preserving emojis in the translation.
    \item Standardizing the translation of slang throughout the corpus.
\end{itemize}

For machine translation, we utilized NLLB-200, a recently released MMNMT model. We employed four variants of the NLLB-200 model for automatically translating the collected dataset: NLLB-200-Distilled 600M, NLLB-200-Distilled 1.3B, NLLB-200 1.3B, and NLLB-200 3.3B. The results of machine translation will be evaluated and analyzed.

\section{Evaluation}
\label{section:6}

An evaluation is conducted to compare the translation results of NLLB across different model sizes. By performing human and machine translations on the corpus, we conducted both automatic and human evaluations to assess the robustness of the NLLB-200 model against naturally occurring noise.

\begin{table*}[t]
    \centering
    \begin{tabular}{lcccc}
        \bottomrule
            \textbf{Evaluation} & \textbf{NLLB-200-Distilled 600M} & \textbf{NLLB-200-Distilled 1.3B} & \textbf{NLLB-200 1.3B} & \textbf{NLLB-200 3.3B}\\ \hline
            BLEU & 11.43 & 10.96 & 12.58 & \textbf{14.04}\\
            CHRF++ & 9.56 & 9.34 & 10.33 & \textbf{11.23}\\
            MQM & 2.21 & 5.54 & 10.70 & \textbf{12.17}\\
        \bottomrule
    \end{tabular}
    \caption{\label{table-4}
    BLEU, CHRF++, and MQM scores of translation results yielded by different models on the dataset.
    }
\end{table*}

\subsection{Evaluation Settings}

For each NLLB model translation results, we conduct the automatic and human evaluation. The source and reference used for the evaluation is from the curated Indonesian-Chinese dataset.

For automatic evaluation, we use automatic metrics of BLEU and CHRF++, which collectively measure both word- and character-level translation quality. We use SacreBLEU tool for calculating the BLEU and CHRF++ score \cite{post-2018-call}.

Additionally, human evaluation is conducted to enhance the evaluation results. To achieve this, we utilize various types of translation errors from the multidimensional quality metric (MQM) framework, which are categorized into five groups: \textit{terminology}, \textit{accuracy}, \textit{fluency}, \textit{local convention}, and \textit{audience appropriateness}. Accuracy encompasses translation errors such as \textit{mistranslation}, \textit{omission}, \textit{addition}, \textit{untranslated}, and \textit{hallucination}. Fluency covers translation errors related to \textit{grammar} and \textit{punctuation}. In NMT, there is a possibility that the NMT system may produce strange or irrelevant translations. Therefore, in our experiment, we introduce a new error type called ``hallucination'', which is not included in the MQM framework. The detailed hierarchy of the error types is presented in Table \ref{table-2}.

Among the ten previously mentioned error types (ET), each one is initially given a standard error weight of 1. However, in the case of hallucination, we assign a weight of 3. This choice is grounded in the recognition that hallucination represents an exceptionally critical error, as it involves a substantial departure in meaning from the source sentence.

Each error type (ET) has three severity levels: \textit{minor}, \textit{major}, and \textit{critical}, with multiplier scores of 1, 5, and 10, respectively. The detail explanation of each severity level is shown in Table \ref{table-3}. We manually identify translation errors for each target translation and annotate the corresponding translation error type and severity level. Additionally, we permit the annotation of multiple translation error types for each translation if different translation errors are found in the target translation.

Following the annotation process, we proceed to calculate the Overall Quality Score (OQS) using the MQM framework. To begin, we calculate the Error Type Penalty Total (ETPT) for each error type, as defined in Equation \ref{eq1}. Subsequently, the OQS is derived by evaluating the relationship between ETPT and the Evaluation Word Count (EWC), as described in Equation \ref{eq2}. The EWC denotes the total count of words present in the source language corpus.

\begin{equation}
    \label{eq1}
    \begin{split}
        \rm ETPT = ({ET}_{{minor }}+ET_{{major }} \times 5 + \\
    \rm ET_{critical } \times 10) \times ET_{{weight }}
    \end{split} 
\end{equation}

\begin{equation}
\label{eq2}
    \rm {OQS} = \left (1-\frac{\sum_{\it{i}}^{}{(ETPT_{\it{i}})}}{EWC}\right )\times 100 
\end{equation}

\begin{table*}[]
\centering
\resizebox{\textwidth}{!}{%
\begin{tabular}{cccccccccccc}
\hline
\textbf{Model Size} &
  \textbf{Error Type} &
  \textbf{Spell/Typo Error} &
  \textbf{Grammar} &
  \textbf{Spoken} &
  \textbf{Slang} &
  \textbf{Proper} &
  \textbf{Dialect} &
  \textbf{Code switch} &
  \textbf{Jargon} &
  \textbf{Emojis} &
  \textbf{Slurs} \\ \hline
\multirow{10}{*}{\textbf{NLLB-200-Distilled 600M}} &
  \textbf{Terminology} &
  6 &
  26 &
  23 &
  16 &
  11 &
  5 &
  5 &
  \textbf{29} &
  2 &
  1 \\
 &
  \textbf{Mistranslation} &
  80 &
  \textbf{396} &
  292 &
  369 &
  172 &
  83 &
  147 &
  85 &
  108 &
  47 \\
 &
  \textbf{Omission} &
  64 &
  \textbf{335} &
  259 &
  317 &
  146 &
  83 &
  137 &
  96 &
  102 &
  34 \\
 &
  \textbf{Addition} &
  6 &
  \textbf{31} &
  24 &
  29 &
  10 &
  5 &
  7 &
  8 &
  6 &
  3 \\
 &
  \textbf{Untranslated} &
  7 &
  \textbf{32} &
  20 &
  28 &
  28 &
  5 &
  12 &
  8 &
  8 &
  1 \\
 &
  \textbf{Hallucination} &
  13 &
  63 &
  58 &
  \textbf{81} &
  23 &
  28 &
  37 &
  11 &
  27 &
  13 \\
 &
  \textbf{Grammar} &
  3 &
  \textbf{38} &
  16 &
  32 &
  15 &
  9 &
  15 &
  8 &
  15 &
  3 \\
 &
  \textbf{Punctuation} &
  18 &
  89 &
  70 &
  \textbf{101} &
  39 &
  26 &
  39 &
  12 &
  30 &
  14 \\
 &
  \textbf{Local Convention} &
  3 &
  \textbf{5} &
  2 &
  4 &
  3 &
  0 &
  0 &
  1 &
  1 &
  0 \\
 &
  \textbf{Audience Appropriateness} &
  1 &
  13 &
  11 &
  \textbf{19} &
  8 &
  3 &
  6 &
  4 &
  3 &
  1 \\ \hline
\multirow{10}{*}{\textbf{NLLB-200-Distilled 1.3B}} &
  \textbf{Terminology} &
  6 &
  18 &
  21 &
  10 &
  1 &
  1 &
  1 &
  \textbf{26} &
  1 &
  0 \\
 &
  \textbf{Mistranslation} &
  64 &
  \textbf{339} &
  255 &
  315 &
  154 &
  82 &
  135 &
  92 &
  101 &
  39 \\
 &
  \textbf{Omission} &
  66 &
  \textbf{349} &
  256 &
  343 &
  152 &
  82 &
  138 &
  92 &
  119 &
  30 \\
 &
  \textbf{Addition} &
  10 &
  47 &
  41 &
  \textbf{49} &
  28 &
  13 &
  17 &
  10 &
  10 &
  5 \\
 &
  \textbf{Untranslated} &
  8 &
  23 &
  14 &
  \textbf{24} &
  19 &
  7 &
  10 &
  9 &
  5 &
  3 \\
 &
  \textbf{Hallucination} &
  14 &
  82 &
  67 &
  \textbf{85} &
  29 &
  22 &
  37 &
  11 &
  25 &
  17 \\
 &
  \textbf{Grammar} &
  6 &
  \textbf{42} &
  25 &
  38 &
  24 &
  10 &
  13 &
  8 &
  12 &
  5 \\
 &
  \textbf{Punctuation} &
  29 &
  \textbf{138} &
  93 &
  131 &
  52 &
  32 &
  50 &
  30 &
  46 &
  15 \\
 &
  \textbf{Local Convention} &
  1 &
  \textbf{3} &
  2 &
  3 &
  2 &
  0 &
  0 &
  3 &
  0 &
  0 \\
 &
  \textbf{Audience Appropriateness} &
  1 &
  11 &
  9 &
  \textbf{15} &
  6 &
  2 &
  4 &
  3 &
  4 &
  0 \\ \hline
\multirow{10}{*}{\textbf{NLLB-200 1.3B}} &
  \textbf{Terminology} &
  5 &
  14 &
  17 &
  7 &
  1 &
  1 &
  1 &
  \textbf{22} &
  1 &
  0 \\
 &
  \textbf{Mistranslation} &
  76 &
  \textbf{361} &
  270 &
  346 &
  178 &
  84 &
  148 &
  92 &
  104 &
  39 \\
 &
  \textbf{Omission} &
  69 &
  \textbf{330} &
  240 &
  295 &
  137 &
  82 &
  138 &
  92 &
  102 &
  32 \\
 &
  \textbf{Addition} &
  4 &
  16 &
  11 &
  \textbf{19} &
  7 &
  6 &
  6 &
  5 &
  3 &
  2 \\
 &
  \textbf{Untranslated} &
  8 &
  \textbf{35} &
  23 &
  32 &
  30 &
  6 &
  15 &
  10 &
  8 &
  4 \\
 &
  \textbf{Hallucination} &
  12 &
  64 &
  65 &
  \textbf{85} &
  21 &
  23 &
  32 &
  14 &
  25 &
  13 \\
 &
  \textbf{Grammar} &
  1 &
  \textbf{46} &
  26 &
  37 &
  21 &
  6 &
  9 &
  12 &
  18 &
  2 \\
 &
  \textbf{Punctuation} &
  25 &
  104 &
  73 &
  \textbf{105} &
  32 &
  34 &
  54 &
  22 &
  35 &
  20 \\
 &
  \textbf{Local Convention} &
  \textbf{2} &
  \textbf{2} &
  1 &
  \textbf{2} &
  \textbf{2} &
  0 &
  0 &
  0 &
  0 &
  0 \\
 &
  \textbf{Audience Appropriateness} &
  1 &
  11 &
  9 &
  \textbf{15} &
  6 &
  1 &
  3 &
  4 &
  4 &
  1 \\ \hline
\multirow{10}{*}{\textbf{NLLB-200 3.3B}} &
  \textbf{Terminology} &
  6 &
  19 &
  20 &
  10 &
  1 &
  1 &
  1 &
  \textbf{27} &
  1 &
  0 \\
 &
  \textbf{Mistranslation} &
  76 &
  \textbf{352} &
  270 &
  344 &
  161 &
  89 &
  155 &
  89 &
  110 &
  43 \\
 &
  \textbf{Omission} &
  70 &
  \textbf{347} &
  251 &
  330 &
  153 &
  81 &
  135 &
  91 &
  103 &
  33 \\
 &
  \textbf{Addition} &
  6 &
  20 &
  24 &
  \textbf{24} &
  12 &
  4 &
  9 &
  4 &
  6 &
  2 \\
 &
  \textbf{Untranslated} &
  11 &
  34 &
  23 &
  \textbf{34} &
  26 &
  8 &
  15 &
  12 &
  9 &
  1 \\
 &
  \textbf{Hallucination} &
  5 &
  60 &
  48 &
  \textbf{67} &
  22 &
  14 &
  23 &
  8 &
  16 &
  12 \\
 &
  \textbf{Grammar} &
  7 &
  \textbf{31} &
  18 &
  22 &
  14 &
  3 &
  7 &
  11 &
  13 &
  4 \\
 &
  \textbf{Punctuation} &
  20 &
  \textbf{88} &
  59 &
  82 &
  41 &
  24 &
  36 &
  15 &
  33 &
  12 \\
 &
  \textbf{Local Convention} &
  0 &
  \textbf{2} &
  \textbf{2} &
  \textbf{2} &
  \textbf{2} &
  0 &
  0 &
  \textbf{2} &
  0 &
  0 \\
 &
  \textbf{Audience Appropriateness} &
  0 &
  8 &
  9 &
  \textbf{14} &
  4 &
  2 &
  3 &
  2 &
  3 &
  1 \\ \hline
\end{tabular}%
}
\caption{The number of translation error types corresponding to different models and noise types on the dataset.}
\label{table-5}
\end{table*}

\begin{figure}[t]
    \centering
    \includegraphics[scale=0.48]{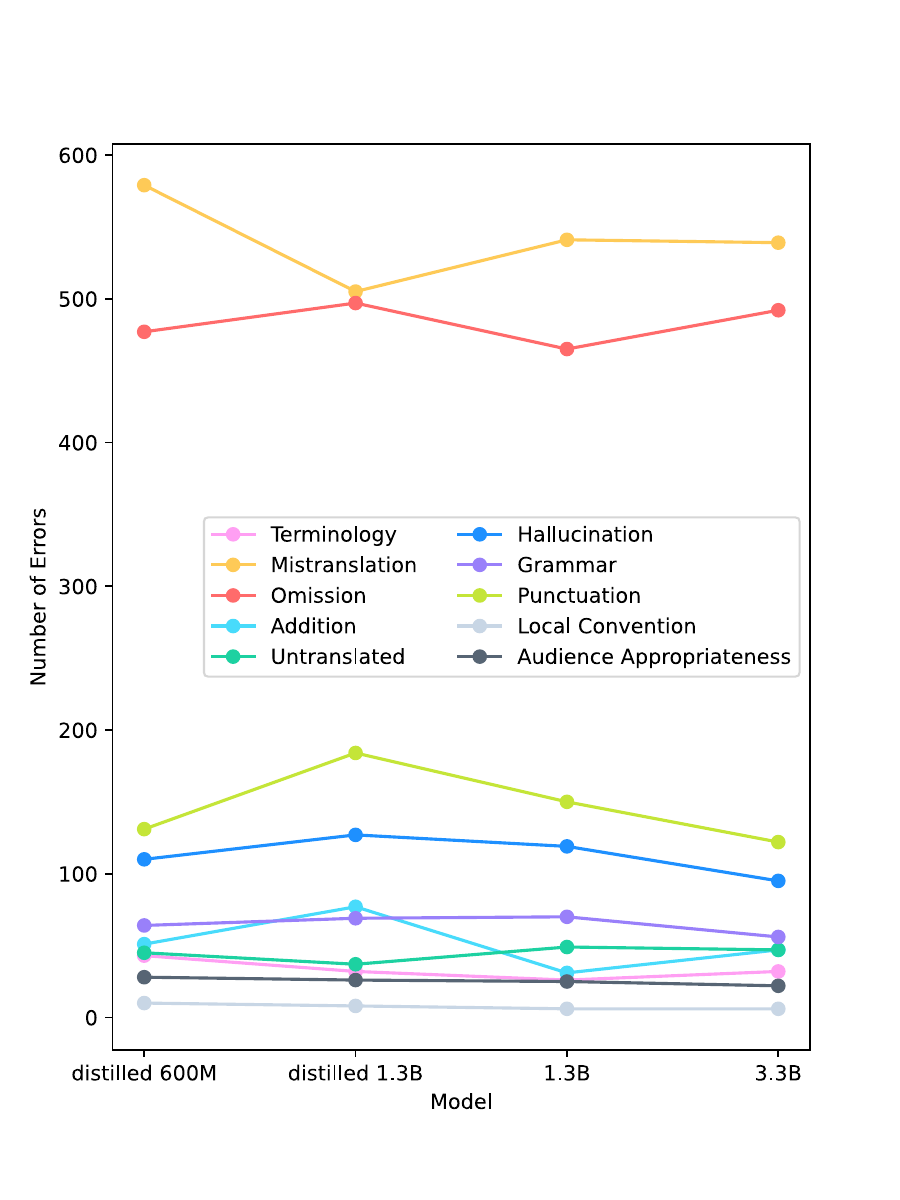}
    \caption{The change of translation error types with the increment of model parameters.}
    \label{fig:2}
\end{figure}

\subsection{Evaluation Results}

First, we conducted an automated evaluation of Indonesian to Chinese translations produced by various models using the BLEU and CHRF++ metrics. The results are represented in Table~\ref{table-4}. Additionally, the results of human evaluation using MQM are also included in Table ~\ref{table-4}. It can be observed that the performance of the models improves as their size increases.

In the human evaluation, Table \ref{table-5} shows the occurrences of translation error types corresponding to different noise types across various models. Based on these findings, we conducted further analysis of the evaluation results in Section \ref{section:7}.

\section{Analysis}
\label{section:7}
\subsection{Effect of Model Size}

Figure \ref{fig:2} illustrates the variation in translation error types as the number of model parameters increases. We conducted analysis for each model size in comparison to the subsequent larger model size. 

In the comparison between distilled 600M and distilled 1.3B model, the number of \textit{mistranslation} errors in the distilled 1.3B model is considerably lower than those in the distilled 600M model. But the number of \textit{omission}, \textit{addition}, and \textit{punctuation} translation error types in the distilled 1.3B model is notably higher compared to the distilled 600M model. We speculate that the model is becoming more proficient in addressing mistranslation. However, it may encounter issues with yielding target translations incompletely or even producing an excessive translation.

Furthermore, when comparing the performance of the distilled 1.3B and 1.3B models, we observed that despite having the same amount of parameters, the 1.3B model exhibits more consistent performance across the 10 translation error types. On the other hand, the distilled 1.3B model demonstrates strong performance in terms of \textit{mistranslation}, but relatively weaker performance in other translation errors. Hence, we conclude that model distillation may enhance the model's proficiency in specific areas while potentially compromising its capability in other areas. Conversely, a model trained directly without distillation may offer a more balanced performance across all areas.

As the model size expands to 3.3B, there is a reduction in the occurence of most error types, with the exception of \textit{mistranslation}, which remains stable, and \textit{addition}, which experiences a substantial increase. This suggests that the increase in model size may lead the MMNMT model to generate additional information.

\subsection{Effect of Noise Types}

\begin{figure}[t]
    \centering
    \includegraphics[scale=0.4]{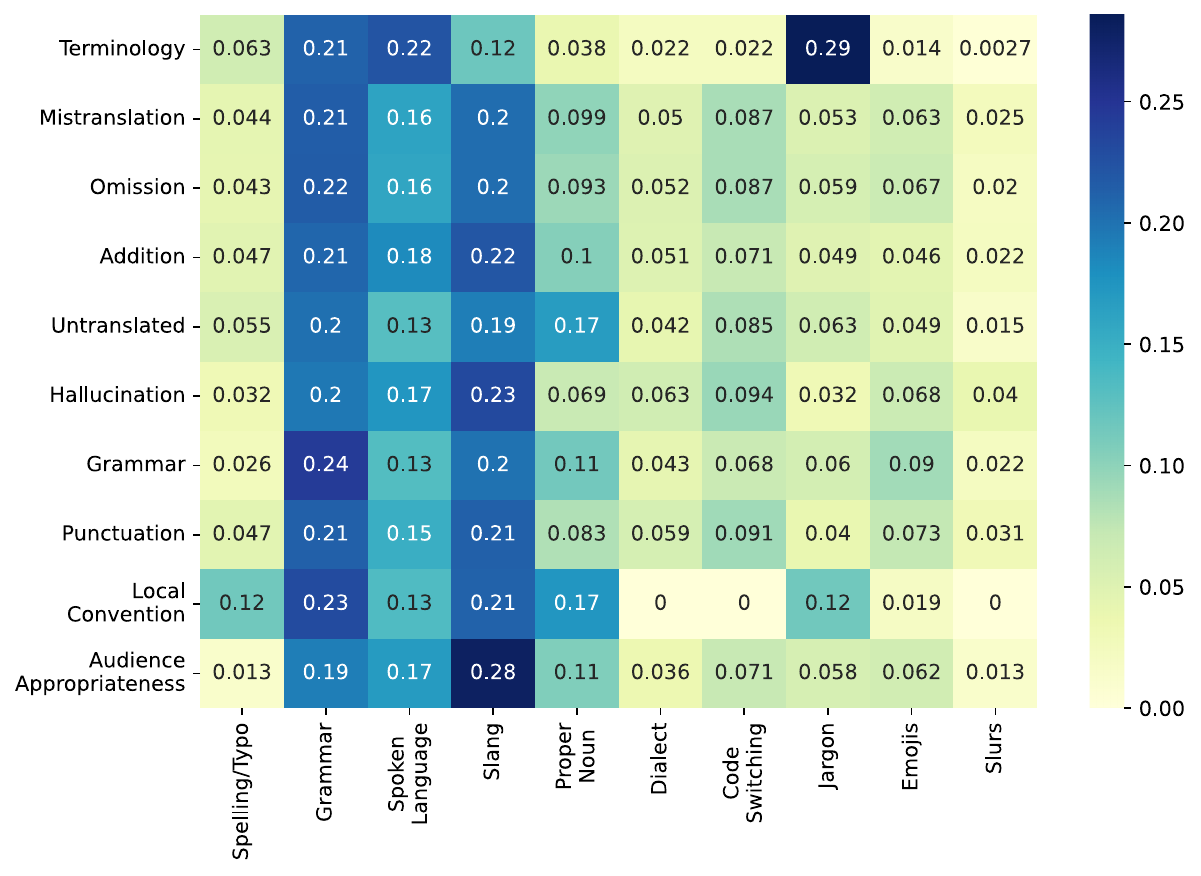}

    \caption{The heatmap of the occurences each of translation errors types according to the noise types.}
    \label{fig:3}
\end{figure}

In Table \ref{table-5}, the occurrence of translation error types based on different noise types is similar across various model sizes. With the exception of \textit{addition}, \textit{untranslated}, and \textit{punctuation}, several models are affected by grammar or slang. However, the differences in counts are not significant. Therefore, we aggregate the results from the 4 models for further analysis.

Figure \ref{fig:3} shows a heatmap that assists in analyzing the occurrence of each translation error type based on the noise types present in the source sentences. The values in the heatmap represent the combined translation results from 4 models, which have been normalized according to each translation error type. The occurrence of \textit{terminology} errors primarily arises from existing jargon in the source sentences. Translation errors related to \textit{accuracy}, \textit{fluency}, and \textit{local convention} primarily arise from slang and grammar noise within the source sentences. Furthermore, errors in \textit{audience appropriateness} predominantly result from slang utilized in the source sentences.



\begin{table}[t]
    \centering
    \begin{tabular}{lcc}
    \bottomrule
        ~ & \textbf{BLEU} & \textbf{CHRF++} \\ \hline
        MQM & 0.8576 & 0.8741 \\ \bottomrule
    \end{tabular}
    \caption{\label{table-6}
Pearson's correlation coefficient between automatically evaluated indicators and human evaluated indicators.
}
\end{table}

\subsection{Relationship between Automatic and Human Evaluation}

It is evident that as the model size increases, both BLEU and CHRF++ scores exhibit a corresponding increase. Moreover, in average, they consistently align with human evaluation results, as measured by MQM. However, there is a slight divergence in the case of the NLLB-200-Distilled-1.3B model size, where there is a slight degradation in both BLEU and CHRF scores. To quantitatively evaluate the correlation between automatic and human assessment metrics, we calculated Pearson's correlation coefficient between the scores obtained from the automatic evaluation indicators and the number of translation errors identified by the human evaluation metrics. The results are presented in Table \ref{table-6}. Based on these findings, we can conclude that CHRF++ demonstrates a stronger correlation with human evaluation when compared to the BLEU score, signifying that CHRF++ serves as a more dependable automatic evaluation metric.

\subsection{Relationship between Sentence Length, Noise Types, and Model Sizes}
\begin{figure}[t]
    \centering
    \includegraphics[scale=0.23]{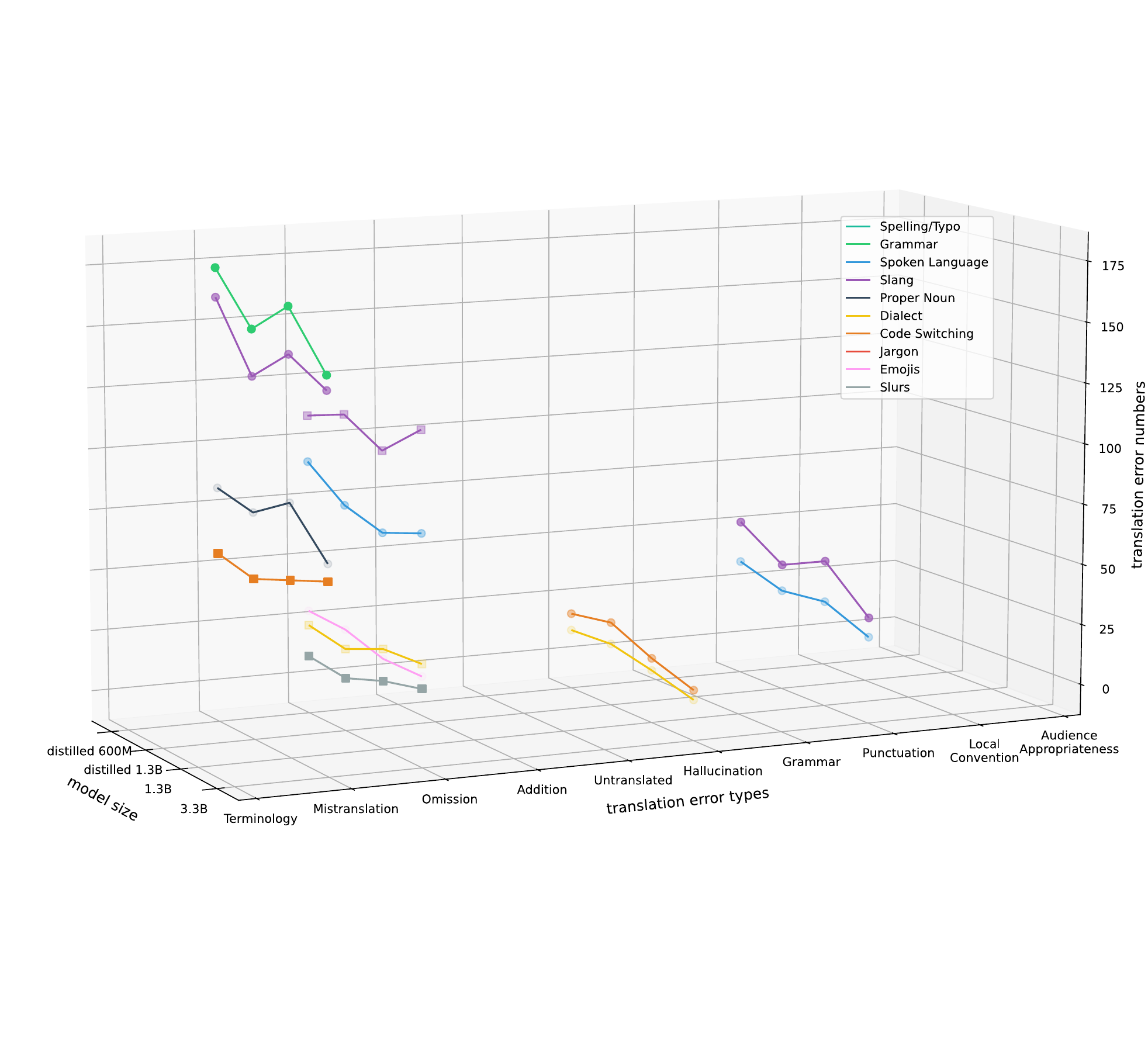}
    \caption{The changing trend of the number of translation errors along with the change of model parameters on short sentences. Dot represents the downward of the trends and square represents the upward of the trends.}
    \label{fig:4}
\end{figure}

\begin{figure}[t]
    \centering
    \includegraphics[scale=0.23]{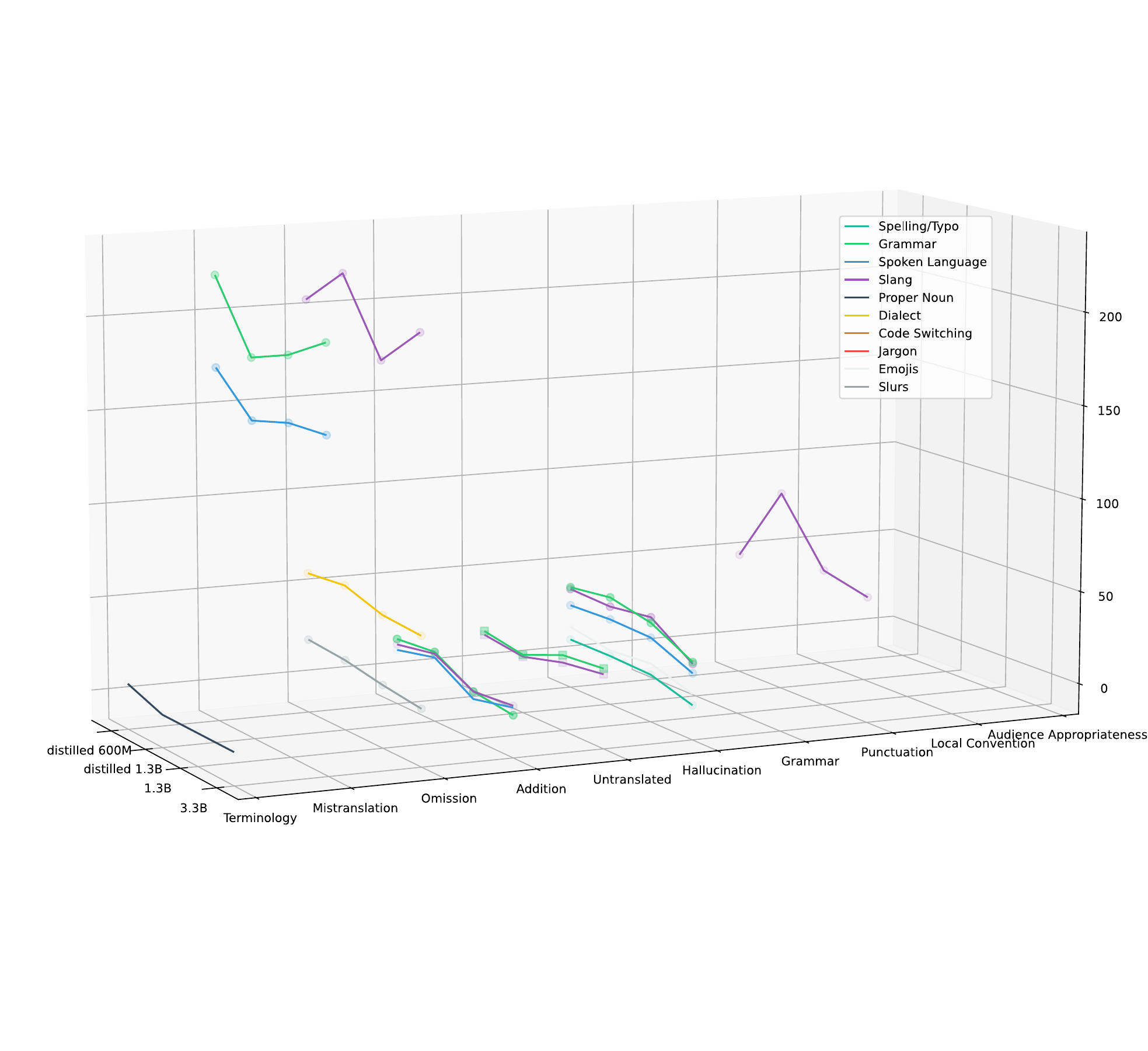}
    \caption{The changing trend of the number of translation errors along with the change of model parameters on long sentences. Dot represents the downward of the trends and square represents the upward of the trends.}
    \label{fig:5}
\end{figure}

When labelling the model translation results with translation error types, we found that there are several differences in the distribution of translation error types for short and long sentences. Thus, we analyzed short and long sentences separately and observed the correlations and differences. We first took the average length of sentences as the threshold value to differentiate long sentences from short sentences. Then we counted the number of translation errors corresponding to different noise types of different models, and fit the number of translation errors generated by different models under different noise types according to the linear regression method. If the primary term coefficient is less than -5 or greater than 0, it means that the number of translation errors has an obvious decreasing or increasing trend as the number of model parameters increases, and we will focus on these cases. 

The results are shown in Figure~\ref{fig:4} and Figure~\ref{fig:5}. The dots on the line represent a clear downward trend in the line, corresponding to a primary term coefficient less than -5. Conversely, the squares represent an upward trend in the line, corresponding to a primary term coefficient greater than 0. The transparency of both the dots and squares indicates the magnitude of these trends, with greater transparency reflecting stronger upward or downward trends.

First, we focus on the analysis of short sentences. In Figure~\ref{fig:4}, it can be observed that as the model size increases, a clear reduction in the number of translation errors related to grammar, slang, spoken language, and proper nouns noise types is evident. Furthermore, there is a noteworthy decline in the number of translation errors attributed to slang and spoken noise types for \textit{punctuation}, indicating an enhancement in fluency. Nevertheless, concerning accuracy, a consistent upward trend is noticeable in the number of translation errors stemming from specific noise types. For instance, \textit{omission} errors associated with slang, dialect, and slurs noise contribute to this ascending trend.

In the case of long sentences, a noticeable pattern emerges as the model size increases. We observed a substantial reduction in the number of translation errors associated with slang, grammar, and spoken language noise types, particularly in terms of accuracy. However, there was a tendency for an increase in \textit{untranslated} translation errors. In terms of fluency, we observed a diminishing trend in \textit{punctuation} errors. Upon comparing these findings, it becomes evident that longer sentences display improved performance with larger models when contrasted with shorter sentences.

\section{Conclusions and Future Work}

The robustness of NMT still poses challenges, especially towards natural occuring noises. Based on research findings, it can be concluded that the size of the model significantly impacts translation performance. In terms of noise types, it is evident that spelling and typographical errors can lead to inaccuracies, fluency issues, and translation errors related to terminology. Larger models perform better on longer sentences.

In the future, we aim to evaluate the robustness of low-resource languages using benchmark datasets. Additionally, given the emergence of large language models (LLMs), we plan to delve into evaluating their performance in translation tasks in comparison to traditional NMT models.

\section*{Limitations}

Our experiments primarily rely on a curated Indonesian-Chinese parallel corpus crawled from Twitter comments with various types of noise. The dataset covers translations only from Indonesian to Chinese and serves as the evaluation benchmark. The dataset size is relatively small for training but is suitable for robustness evaluation. Due to limitation in computational resources, the NLLB-200 54B model is not used in this research.

\section*{Ethics Statement}
\paragraph{Data Privacy}
The curated noisy parallel corpus of Indonesian-Chinese is openly available for research purposes. One concern regarding this data is that it is obtained by crawling Twitter comments, raising privacy concerns for Twitter users. In order to protect the privacy of Twitter users, we have removed user IDs and usernames from the dataset. In Twitter comments, it is common for users to tag other users by their usernames. To address this, we have systematically removed any text that includes the "@" prefix, thereby effectively eliminating tagged usernames.

\paragraph{Social Impact}
However, it is essential to note that the dataset is sourced from social media comments, where certain comments may not be appropriate for all audiences, including those containing hate speech or offensive language. In our experiments, we maintain these comments for the purpose of robustness evaluation, as they are also considered as a form of natural noise. It is crucial to take this aspect into account before utilizing our parallel corpus for other studies.

\section*{Acknowledgments}
The present research was supported by the Natural Science Foundation of Xinjiang Uygur Autonomous Region (No. 2022D01D43). We would like to appreciate the insightful comments provided by anonymous reviewers. We extend our gratitude to Ms. Yu for her assistance in proofreading the translation of the parallel corpus. Additionally, we gratitudely acknowledge the support of the Chinese Government Scholarship (CGS).

\nocite{*}
\section{Bibliographical References}\label{sec:reference}

\bibliographystyle{lrec-coling2024-natbib}
\bibliography{lrec-coling2024-example}


\end{document}